\pdfoutput=1

\documentclass[11pt]{article}

\usepackage{acl}

\usepackage{times}
\usepackage{latexsym}

\usepackage[T1]{fontenc}

\usepackage[utf8]{inputenc}

\usepackage{microtype}

\usepackage{inconsolata}

\usepackage{multirow}
\usepackage{graphicx}
\usepackage{amsmath,amsfonts,amssymb}
\usepackage{subcaption}
\usepackage{array}
\usepackage{makecell}
\usepackage{multicol}
\usepackage{booktabs}

\title{Enhancing Contextual Understanding in Large Language Models \\through Contrastive Decoding}

\author{Zheng Zhao\textsuperscript{1,\footnotemark} \quad Emilio Monti\textsuperscript{2} \quad Jens Lehmann\textsuperscript{2} \quad Haytham Assem\textsuperscript{2} \\
  \textsuperscript{1}School of Informatics, University of Edinburgh \\
  \textsuperscript{2}Amazon \\
  \texttt{zheng.zhao@ed.ac.uk},
  \texttt{\{monti,jlehmnn,hithsala\}@amazon.com}}

\begin{document}
\maketitle
\def\thefootnote{*}\footnotetext{Work done during an internship at Amazon.}\def\thefootnote{\arabic{footnote}}
\begin{abstract}
Large language models (LLMs) tend to inadequately integrate input context during text generation, relying excessively on encoded prior knowledge in model parameters, potentially resulting in generated text with factual inconsistencies or contextually unfaithful content. LLMs utilize two primary knowledge sources: 1) prior (parametric) knowledge from pretraining, and 2) contextual (non-parametric) knowledge from input prompts. The study addresses the open question of how LLMs effectively balance these knowledge sources during the generation process, specifically in the context of open-domain question answering. To address this issue, we introduce a novel approach integrating contrastive decoding with adversarial irrelevant passages as negative samples to enhance robust context grounding during generation. Notably, our method operates at inference time without requiring further training. We conduct comprehensive experiments to demonstrate its applicability and effectiveness, providing empirical evidence showcasing its superiority over existing methodologies. Our code is publicly available.\footnote{\url{https://github.com/amazon-science/ContextualUnderstanding-ContrastiveDecoding}}
\end{abstract}

\section{Introduction}
Improving large language models (LLMs) has been a primary focus in natural language processing research. Recent strides have incorporated retrieval mechanisms to enhance LLMs \cite{lewis-etal-2020-retrieval,guu-etal-2020-realm,izacard-grave-2021-leveraging,izacard-etal-2023-atlas}, augmenting their ability to produce contextually relevant and precise responses \cite{min-etal-2023-nonparametric, mallen-etal-2023-trust}. Retrieval-augmented LLMs, which leverage both \textit{parametric} knowledge acquired during training and \textit{non-parametric} knowledge retrieved during inference, exhibit potential in addressing challenges such as limited memorization \cite{kandpal-etal-2023-large}, knowledge conflicts \cite{longpre-etal-2021-entity}, and outdated information \cite{kasai2023realtime}.

\begin{figure}
    \centering
    \includegraphics[width=\linewidth]{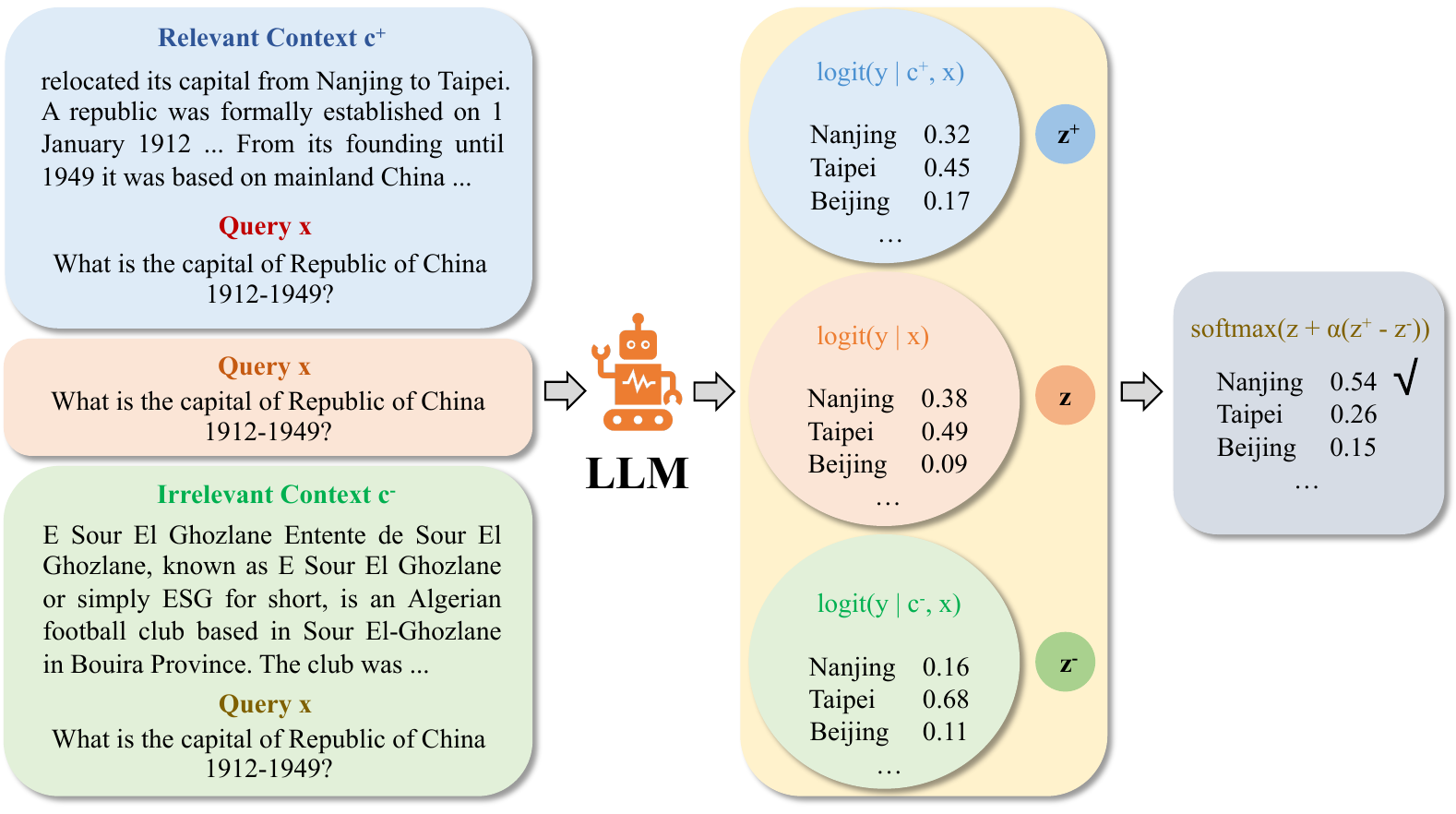}
\caption{\label{fig:method}An illustration of our proposed decoding method. Despite the relevant context suggesting the answer as ``Nanjing'', it contradicts the LLM's prior knowledge. After reconciling different knowledge sources, the model correctly predicted the answer by boosting Nanjing's plausibility and reducing Taipei's likelihood. This decision was based on considering Nanjing to be less likely given the irrelevant context, while Taipei is deemed more probable.}
\vspace{-1ex}
\end{figure}

An ongoing question pertains to how LLMs ought to balance these two knowledge sources during generation. Previous research suggests that LLMs can falter in adequately attending to newly introduced information within the contextual knowledge. To tackle this issue, context-aware decoding (CAD; \citealp[]{shi2023trusting}) has been proposed. By employing a contrastive output distribution, CAD highlights discrepancies in output probabilities when the model operates with and without context. Their experiments illustrate CAD's effectiveness in overriding the model's parametric knowledge in cases of conflict with provided context. However, while prior works often assert context as inherently reliable, our perspective argues that LLMs should possess the capacity to navigate and reconcile both parametric and non-parametric knowledge, ultimately refining their ability to strike a judicious balance. This paper undertakes the development and assessment of a novel decoding strategy tailored for retrieval-augmented LLMs, seeking equilibrium in utilizing parametric and non-parametric knowledge sources. The proposed method involves a contrastive decoding approach \cite{li-etal-2023-contrastive}, integrating both relevant and irrelevant contexts, wherein the irrelevant context can be adversarially crafted retrieval or bottom-ranked retrieved text. Notably, we emphasize the criticality of leveraging irrelevant contexts, a distinguishing feature of our approach, with the expectation that the model will diverge from incorrect responses.

We conducted extensive experiments on diverse datasets like Natural Questions \cite{kwiatkowski-etal-2019-natural}, TriviaQA \cite{joshi-etal-2017-triviaqa}, and PopQA \cite{mallen-etal-2023-trust}. We employed a range of vanilla LLMs, including OPT \cite{zhang2022opt}, Falcon \cite{almazrouei2023falcon}, LLaMA families \cite{touvron2023llama1,touvron2023llama2}, and instruction-tuned Flan-T5 \cite{chung2022scaling}. Through this comprehensive evaluation, we provide empirical evidence supporting the superiority of incorporating irrelevant contexts in assisting LLMs to manage knowledge conflicts and seamlessly integrate contexts for generating responses in open-domain question answering against conventional decoding approaches without necessitating further fine-tuning. We also explore the impact of different retrieval sources on the decoding strategy, emphasizing the importance of refining retrieval mechanisms for further enhancements in performance.

Additionally, the paper explores different facets of the proposed decoding approach, including the impact of various hyperparameters, the effect of scaling model sizes, and the selection of irrelevant contexts. This exploration provides deeper insights into leveraging parametric and non-parametric knowledge sources. We demonstrate that although our approach outperforms regular decoding across most model sizes, it particularly excels with larger models. Moreover, we show our method's effectiveness even with simple fixed irrelevant contexts. Additionally, our approach exhibits consistent performance improvements in answering questions with knowledge across varying levels of popularity. Beyond benchmarking against existing methods, this study also explores practical implications and constraints of the proposed decoding strategy, delineating pathways for future research in generative tasks beyond question answering.

\section{Related Works}
\paragraph{Retrieval-augmented LLMs}
While LLMs relying solely on their parameters can capture extensive world knowledge, they exhibit limited memorization for less frequent entities \cite{kandpal-etal-2023-large}, susceptibility to hallucinations \cite{shuster-etal-2021-retrieval-augmentation}, and temporal degradation \cite{luu-etal-2022-time,jang-etal-2022-temporalwiki}. Furthermore, the acquired parametric knowledge swiftly becomes outdated \cite{kasai2023realtime}. Recent research emphasizes the enhancement of LLMs with non-parametric memories, referred to as retrieved text chunks, enabling smaller models to match the performance of larger counterparts \cite{izacard-etal-2023-atlas}. Studies exploring the integration of retrieved non-parametric memories within intermediate states or output spaces have shown effectiveness in overcoming LLM limitations in memorization and knowledge updating \cite{zhong-etal-2022-training,min-etal-2023-nonparametric}. \citet{mallen-etal-2023-trust} extensively analyze the circumstances favoring the benefits of retrieval augmentation. They demonstrate its efficacy in less frequent occurrences but caution about potential misguidance for LLMs. Building upon these insights, they introduce adaptive retrieval and empirically showcase its promising effectiveness.

\paragraph{Knowledge Conflicts}
In cases of conflicting knowledge in updated documents, language models are expected to generate responses based on provided contexts rather than relying solely on outdated parametric knowledge. Retrieval-augmented LLMs \cite{min-etal-2023-nonparametric,shi2023replug,izacard-etal-2023-atlas} particularly benefit from this scenario by employing externally retrieved documents to enrich their knowledge. However, the mere addition of documents doesn't consistently influence model predictions, as current LLMs often overlook contexts and heavily rely on prior parametric knowledge \cite{longpre-etal-2021-entity,chen-etal-2022-rich}. \citet{zhou-etal-2023-context} aim to improve a model's fidelity to context using prompting-based method, but are constrained to large-scale instruction-finetuned LLMs like OpenAI’s gpt-3.5-turbo-instruct. \citet{zhang-etal-2023-merging} address how to combine retrieved and parametric knowledge to get the best of both worlds for open-domain QA, but their method requires further training discriminators with silver labels. In contrast, our work investigates a decoding strategy applicable to any LLMs without any training.

\paragraph{Contrastive Decoding}
The exploration of contrastive decoding methods extensively addresses text generation. MMI-based decoding \cite{li-etal-2016-diversity} utilizes a contrastive formulation to enhance output diversity in dialog generation. DExperts \cite{liu-etal-2021-dexperts} dampens the output distribution of an anti-expert (e.g., exposed to toxic language) to guide generations away from undesired attributes. Contrastive decoding \cite{li-etal-2023-contrastive} demotes an amateur model (e.g., models with minimal parameters) to distill expert knowledge from larger, competitive models. \citet{pozzobon-etal-2023-goodtriever} introduce an innovative toxicity mitigation approach that contrasts and ensembles the next token probabilities obtained from a LLM using both toxic and non-toxic retrievals. Context-aware decoding \cite{shi2023trusting} emphasizes output probability differences using a contrastive ensemble between model predictions with and without non-parametric knowledge. It effectively overrides a model's parametric knowledge when it conflicts with the provided non-parametric information. While our work builds upon the concept of context-aware decoding, one key distinction lies in the integration of irrelevant context. Unlike \citeauthor{shi2023trusting}'s approach, which focuses solely on relevant non-parametric knowledge, our method incorporates potentially irrelevant non-parametric knowledge into the inference process with the expectation that the model will deviate from incorrect responses.

\section{Methodology}
\subsection{Problem Statement}
We consider decoding approaches for open-domain question answering, where the large language model $\theta$ receives an input query $\boldsymbol{x}$ and aim to generate a faithful answer $\boldsymbol{y}$. 
During the generation of $\boldsymbol{y}_t$ at each time step $t$, the language model computes the logits $\boldsymbol{\mathrm{z}}_t \in \mathbb{R}^{|V|}$ for the $t$-th token, where $V$ represents the vocabulary. 
The probability distribution over the vocabulary is derived by normalizing and exponentiating $\boldsymbol{z}_t$ as follows:
\begin{equation*}
    p_{\theta}(y_t | \boldsymbol{x},\boldsymbol{y}_{<t}) = \mathrm{softmax}(\boldsymbol{\mathrm{z}}_t)\text{.}
\end{equation*}

Prompting the model for its parametric knowledge involves sampling the response from the probability distribution conditioned on the query $\boldsymbol{x}$ and the previously generated response $\boldsymbol{y}_{<t}$:

\begin{equation*}
    y_t \sim  p_{\theta}(y_t | \boldsymbol{x},\boldsymbol{y}_{<t})\text{.}
\end{equation*}

Similarly, when incorporating additional context $\boldsymbol{c}$, containing external knowledge beyond the model’s parametric knowledge, our model $\theta$ generates a response $\boldsymbol{y}$ considering the query, context, and the previously generated response:

\begin{equation*}
y_t \sim  p_{\theta}(y_t | \boldsymbol{c},\boldsymbol{x},\boldsymbol{y}_{<t})\text{.}
\end{equation*}

We observe two sources of knowledge (parametric vs. non-parametric) contributing to model responses, which may sometimes conflict \cite{longpre-etal-2021-entity,neeman-etal-2023-disentqa}. While some argue for prioritizing non-parametric knowledge over potentially outdated parametric knowledge \cite{shi2023trusting}, we propose the importance of striking a balance between these sources as non-parametric knowledge, derived from external retrievers, may also contain inaccuracies.

\subsection{Multi-Input Contrastive Decoding}
Context can be both beneficial and problematic. Thus, we segregate context $\boldsymbol{c}$ into relevant $\boldsymbol{c}^+$ and irrelevant $\boldsymbol{c}^-$. At each decoding time step $t$, our approach combines the model's prediction based on its parametric knowledge ($\boldsymbol{\mathrm{z}}_t$) with predictions utilizing relevant ($\boldsymbol{\mathrm{z}}_{t}^{+}$) and irrelevant ($\boldsymbol{\mathrm{z}}_{t}^{-}$) contexts:

\begin{equation*}
y_t \sim \mathrm{softmax}(\boldsymbol{\mathrm{z}}_t + \alpha (\boldsymbol{\mathrm{z}}_{t}^{+} - \boldsymbol{\mathrm{z}}_{t}^{-}))\text{,}
\end{equation*}

\noindent where $\alpha$ is a hyperparameter that governs the extent of modification to the parametric answer ($\boldsymbol{\mathrm{z}}_t$). Equivalently, 

\begin{align*}
    y_t &\sim \tilde p_{\theta}(y_t | \boldsymbol{c}^+,\boldsymbol{c}^-,\boldsymbol{x},\boldsymbol{y}_{<t}) \\
    &\propto p_{\theta}(y_t | \boldsymbol{x},\boldsymbol{y}_{<t})\left( \frac{ p_{\theta}(y_t | \boldsymbol{c}^{+},\boldsymbol{x},\boldsymbol{y}_{<t})}{ p_{\theta}(y_t | \boldsymbol{c}^{-},\boldsymbol{x},\boldsymbol{y}_{<t})}\right)^{\alpha}\text{.}
\end{align*}

In essence, a response will exhibit high probability only if it holds high likelihood under both learned parametric knowledge and relevant non-parametric knowledge, while demonstrating low probability under irrelevant non-parametric knowledge. The ratio  $\frac{ p_{\theta}(y_t | \boldsymbol{c}^{+},\boldsymbol{x},\boldsymbol{y}_{<t})}{ p_{\theta}(y_t | \boldsymbol{c}^{-},\boldsymbol{x},\boldsymbol{y}_{<t})}$ functions as a scaling factor used to modify the parametric answer for the given input query. A larger $\alpha$ implies a greater modification, with $\alpha=0$ resulting in no modification, indicating regular decoding using solely parametric knowledge without additional context.

Fundamentally, our proposed decoding operates as an ensemble involving the logits $\boldsymbol{\mathrm{z}}_t$, $\boldsymbol{\mathrm{z}}_t^+$, and $\boldsymbol{\mathrm{z}}_t^-$. A similar ensemble approach has been explored in \citet{liu-etal-2021-dexperts} and \citet{li-etal-2023-contrastive} for controllable and open-ended text generation, though their ensembles are based on predictions from different models. Another similar work to ours is CAD \cite{shi2023trusting}, which examines scenarios where the model's parametric knowledge contradicts non-parametric knowledge. CAD essentially constitutes a contrastive ensemble between $\boldsymbol{\mathrm{z}}_t$ and $\boldsymbol{\mathrm{z}}_t^+$. In this study, we concentrate on the general case of open-domain question answering, proposing a dynamic adjustment of $\alpha$, controlling the degree of modification without treating it as a fixed hyperparameter. We provide an illustration of our method in Figure~\ref{fig:method}.

\paragraph{Dynamic $\alpha$}
In prior logit adjustment methods \cite{liu-etal-2021-dexperts,malkin-etal-2022-coherence,oneill2023steering,shi2023trusting,pozzobon-etal-2023-goodtriever}, $\alpha$ remains a fixed hyperparameter, requiring exhaustive search within the parameter space. Our innovation lies in dynamically setting $\alpha$ at each time step $t$ without supervision, enabling fine-grained token-level adjustments. We estimate LLM confidence following \citet{jiang-etal-2021-know} by computing the highest probability from the normalized predicted token probabilities at each step:

\begin{equation*}
    C = \max_{y' \in V} P_{\theta}(y' | \boldsymbol{x},\boldsymbol{y}_{<t})\text{.}
\end{equation*}

Similarly, we estimate LLM confidence using relevant non-parametric knowledge $\boldsymbol{c}^+$:

\begin{equation*}
    C_{R} = \max_{y' \in V} P_{\theta}(y' | \boldsymbol{c}^+,\boldsymbol{x},\boldsymbol{y}_{<t})\text{.}
\end{equation*}

At each time step, the value of $\alpha$ is determined as follows:

\begin{equation*}
    \alpha = 
    \begin{cases}
    1 - C,& \text{if } C > C_R,\\
    C_R,              & \text{otherwise.}
    \end{cases}
\end{equation*}

Our rationale is that higher LLM confidence in parametric knowledge warrants minor adjustments, while greater confidence in relevant non-parametric knowledge necessitates more substantial modifications to the parametric answer. Note that we use $1 - C$ instead of using $C - C_R$ to avoid the case where both $C$ and $C_R$ are low. In such case, a larger modification is still desired.

\paragraph{Selection of $\boldsymbol{c}^+$ and $\boldsymbol{c}^-$}
Choosing relevant context $\boldsymbol{c}^+$ is straightforward and we follow the retrieval-augmented LLM literature where we use the top retrieved texts from a retrieval module by running our input query over an external knowledge base.
However, selecting irrelevant context $\boldsymbol{c}^-$ is not trivial. 
Potential methods include using lower-ranked retrievals, random text, or even deliberately crafted adversarial text. 
The primary aim of $\boldsymbol{c}^-$ is to provide adversarial knowledge to elicit incorrect predictions that can be disregarded from the final token distribution.
We explore various strategies for selecting $\boldsymbol{c}^-$ in Section~\ref{sec:effect-of-irrelevant-c}.

\section{Experimental Setup}
The present study revolves around open-domain question answering, which involves tasking models to generate responses to factual questions in natural language. Specifically, we concentrate on the \textit{open-book} QA setting \cite{roberts-etal-2020-much}, where we harness non-parametric knowledge by supplying relevant contexts along with the question itself to the model during inference. Consistent with prior investigations, we utilize prompting techniques to assess the models' performance.

\subsection{Datasets and Metrics}
\paragraph{Datasets} 
Our method undergoes evaluation using three popular QA benchmarks: TriviaQA \cite{joshi-etal-2017-triviaqa}, Natural Questions (NQ; \citealt{kwiatkowski-etal-2019-natural}), and PopQA \cite{mallen-etal-2023-trust}. TriviaQA comprises trivia questions sourced from the Web, whereas NQ consists of questions derived from actual Google search queries, with answer spans located in Wikipedia articles identified by annotators. PopQA is a novel entity-centric open-domain QA dataset covering factual information about entities across a spectrum of popularity, including \textit{long-tail} knowledge often overlooked in other popular QA datasets.

\paragraph{Metrics} 
In line with prior research, our primary metric for evaluating performance is the exact match (EM), which determines whether the predicted sequence matches precisely with one of the correct answers provided within the dataset.

\subsection{Baselines and Models}
\paragraph{Baselines} 
Baseline approaches include regular decoding with greedy decoding, following prior work \cite{izacard-grave-2021-leveraging}. We prompt the model for an answer by providing contextual information. While our primary focus remains on the \textit{open-book} QA setting, we also present a baseline employing the \textit{closed-book} QA setting, where the prompt consists solely of questions. This exploration aims to scrutinize the parametric knowledge of LLM. Additionally, we compare our method to CAD, which accentuates the difference in output probabilities when employing a model with and without context.

\paragraph{Models} 
Our decoding method undergoes evaluation across models varying in scale: Flan-T5 (XL-3B, XXL-11B; \citealt{chung2022scaling}), Falcon (7B, 40B; \citealt{almazrouei2023falcon}), OPT (6.7B, 13B, 30B, 66B; \citealt{zhang2022opt}), Llama (7B, 13B, 33B, 65B; \citealt{touvron2023llama1}), and Llama 2 (7B, 13B, 70B; \citealt{touvron2023llama2}), without additional fine-tuning.

\paragraph{Instructions} 
We employ a straightforward template, i.e., ``{\fontfamily{lmtt}\selectfont Answer the following question. Question: <question> Answer:}'', to format all questions for generative prediction in the closed-book setting. For the open-book setting, the template becomes ``{\fontfamily{lmtt}\selectfont Answer the question based on the context below. Context: <context> Question: <question> Answer:}''. Although more sophisticated prompts were trialed in preliminary experiments, their marginal improvement over the simple template did not warrant their use, especially considering the risk of overfitting the model. In alignment with prior work \cite{chung2022scaling}, we employ 5-shot prompting for all models.

\paragraph{Retrieval models}
As previously mentioned, we explore a retrieval-augmented LLM approach in the open-book setting. This involves running an off-the-shelf retrieval system offline to obtain relevant context from Wikipedia for each query\footnote{We utilize the Wikipedia dump from 2018.}, which is then concatenated with the original query. We utilize two widely-used retrieval systems: BM25 \cite{robertson-hugo-2009-bm25} and Contriever \cite{izacard-etal-2022-unsupervised}. BM25 operates as a static term-based retriever without training, while Contriever is pre-trained on extensive unlabeled corpora. In this study, we leverage Contriever-MS MARCO, a Contriever fine-tuned on MS MARCO \cite{bajaj2018ms}. Consistent with \citet{mallen-etal-2023-trust}, we utilize the top one retrieved paragraph. Additionally, TriviaQA and NQ datasets provide gold contexts, which we employ to measure the theoretical upper bound of our proposed decoding method. We also investigate the impact of using different retrieval methods in Section~\ref{sec:effect-of-context}.

\begin{table}[ht]
\small
\centering
\begin{tabular}{llccc}
\toprule
{\textbf{Model}}  & \textbf{Decoding} & {\textbf{NQ}} &{\textbf{TQA}} & {\textbf{PopQA}} \\
                           \midrule
\multirow{5}{*}{Flan-T5 11B} & Reg.-Cl.  & 14.82  &   40.5  &  13.98     \\
                            & Reg.-Op.  & 57.84 & 79.36 & 31.16      \\
                            & CAD      & 47.56 & 66.08 &  26.28     \\
                            & Ours-F    & 59.58 & 76.75  & 31.37   \\  
                            & Ours-D    & \textbf{63.16} & \textbf{80.09}  & \textbf{34.64}   \\ 
\midrule
\multirow{5}{*}{Falcon 40B} & Reg.-Cl.  & 28.56 &  71.74   &  28.79    \\
                         & Reg.-Op.  &  53.32 &  72.05  &   39.16    \\
                         & CAD      & 49.36 & 20.72   &  35.31     \\
                         & Ours-F     & \textbf{53.77} & 79.56  & \textbf{39.87}   \\  
                         & Ours-D     & 50.53 & \textbf{80.73} & 38.28   \\ 
\midrule
\multirow{5}{*}{OPT 66B} & Reg.-Cl.  & 13.71 & 39.65 & 15.62  \\
                         & Reg.-Op.  &  48.73 & 62.38 & 34.77  \\
                         & CAD      & 45.93  &  24.51   & 33.45      \\
                         & Ours-F     & \textbf{51.97}  &  \textbf{68.11} &  \textbf{34.83}  \\  
                         & Ours-D     & 44.41 & 63.89  &  33.44  \\ 
\midrule
\multirow{5}{*}{Llama 65B} & Reg.-Cl.  & 34.13 &  75.72   &  35.9    \\
                        & Reg.-Op.  &  55.32  &  74.76   &   40.31    \\
                        & CAD      &  48.03  & 24.51    &  31.97     \\
                         & Ours-F     & \textbf{57.01} & 76.61  & 39.9  \\  
                         & Ours-D     & 52.35 & \textbf{80.28}  &  \textbf{40.58}  \\ 
\midrule
\multirow{5}{*}{Llama-2 70B} & Reg.-Cl.  &  37.87 &  79.69   &   40.98    \\
                           & Reg.-Op.  & 56.07  &  76.07  &  42.7   \\
                           & CAD      & 47.53  &  31.36   &   33.05    \\
                         & Ours-F     & \textbf{58.86} & 78.38  & 42.59   \\  
                         & Ours-D     & 55.24 & \textbf{81.7}  &  \textbf{44.3}  \\ 

\bottomrule
\end{tabular}
\caption{Results of models using gold retrieval (NQ, TriviaQA), and Contriever retrieval (PopQA). Reg.-Cl. refers to regular decoding with \textit{closed-book} setting (i.e. no retrieval). Reg.-Op. refers to regular decoding with \textit{open-book} setting (i.e. with retrieval). Ours-F refers to our method utilizing a fixed alpha, while Ours-D designates our method incorporating a dynamic alpha.}
\label{tab:main_results}
\end{table}

\paragraph{Setting alpha}
Our approach introduces a hyperparameter $\alpha$ to govern the degree of modification atop LLM's parametric knowledge. For CAD, after a grid search using the validation set, we set $\alpha = 0.5$. In fixed alpha experiments for our method, we set $\alpha = 1.0$. In dynamic alpha experiments, we do not have to set alpha values explicitly. We explore the effect of $\alpha$ on our method in Section~\ref{sec:effect-of-alpha}.

\begin{figure*}[t]
    \centering
    \includegraphics[width=\linewidth]{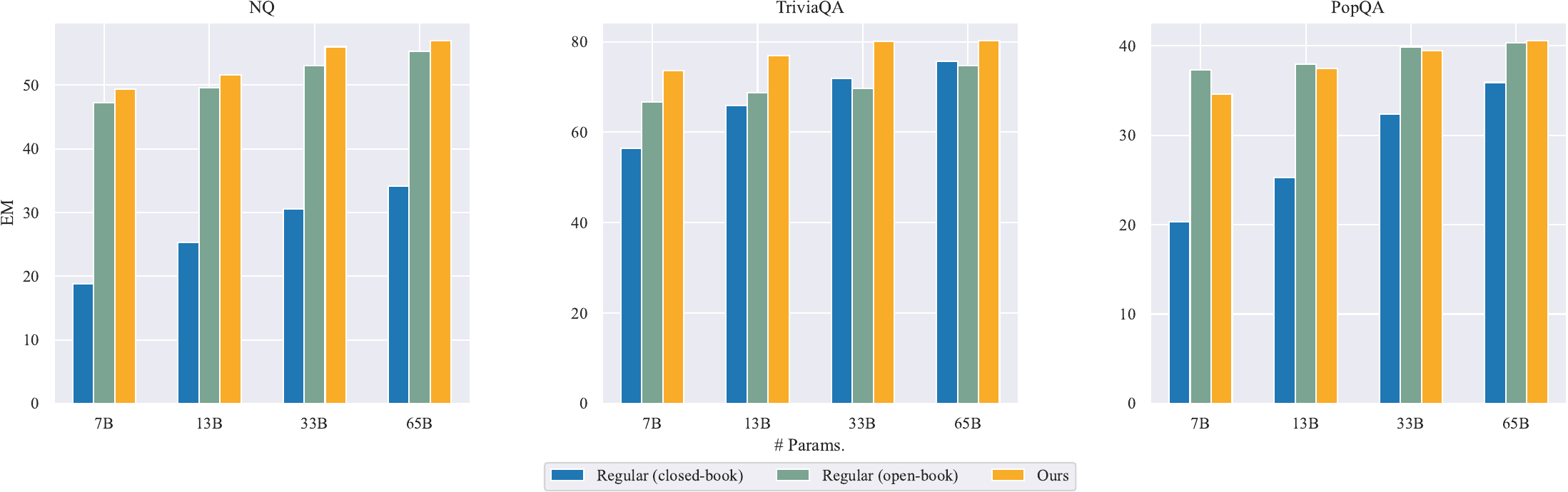}
\caption{\label{fig:scaling-effect-llama} Performance comparison of our method against regular decoding across various sizes of Llama 1 models.}
\end{figure*}

\section{Results}
We present the results of models featuring the largest variants in Table~\ref{tab:main_results}. Notably, employing regular decoding within an open-book setting consistently outperforms the closed-book setting across most models. This inclination suggests that LLM systems require non-parametric knowledge to excel in tasks demanding substantial knowledge assimilation. Interestingly, the performance of Llama 65B and Llama 2 70B in the closed-book setting surpasses that in the open-book setting concerning TriviaQA, indicating these models' proficiency in factual knowledge retention without resorting to non-parametric knowledge. This finding possibly implies that TriviaQA, being the oldest dataset among the three, potentially overlaps with the training data of these LLMs.

Crucially, our proposed decoding approach demonstrates superior performance across all three datasets compared to both regular decoding and CAD.\footnote{The CAD results were based on our implementation, due to the unavailability of the original CAD implementation at the time of our study.} Noteworthy variations exist in the efficacy of employing either the fixed alpha strategy or the dynamic alpha strategy; while in certain instances the fixed alpha approach exhibits better performance, the dynamic alpha approach outperforms in others. In subsequent references within this paper, when mentioning our method, we refer to the setting that delivers superior performance based on Table~\ref{tab:main_results}, without explicitly specifying whether it involves dynamic or fixed alpha.

\subsection{Effect of Model Scaling}

Thus far, our study has elucidated the efficacy of our proposed decoding approach across diverse model families. This segment aims to examine the impact of scaling the model's parameter count on our methods. The results pertaining to Llama variants--specifically, Llama 7B, 13B, 33B, and 65B--are illustrated in Figure~\ref{fig:scaling-effect-llama}. We provide the results of scaling for other models in Appendix~\ref{app:scaling_results}. An observable trend emerges wherein, with an increase in model size, the disparity between closed-book and open-book performance diminishes, indicating that larger models possess greater potential for assimilating parametric knowledge. Furthermore, our decoding method consistently outperforms regular decoding across all model sizes, except for a few instances in the case of PopQA with smaller model variants. We posit this discrepancy to the absence of gold context within the PopQA dataset, leading to reliance on Contriever's retrieval, which may occasionally introduce inaccuracies.

\begin{figure*}[ht]
    \centering
    \includegraphics[width=\linewidth]{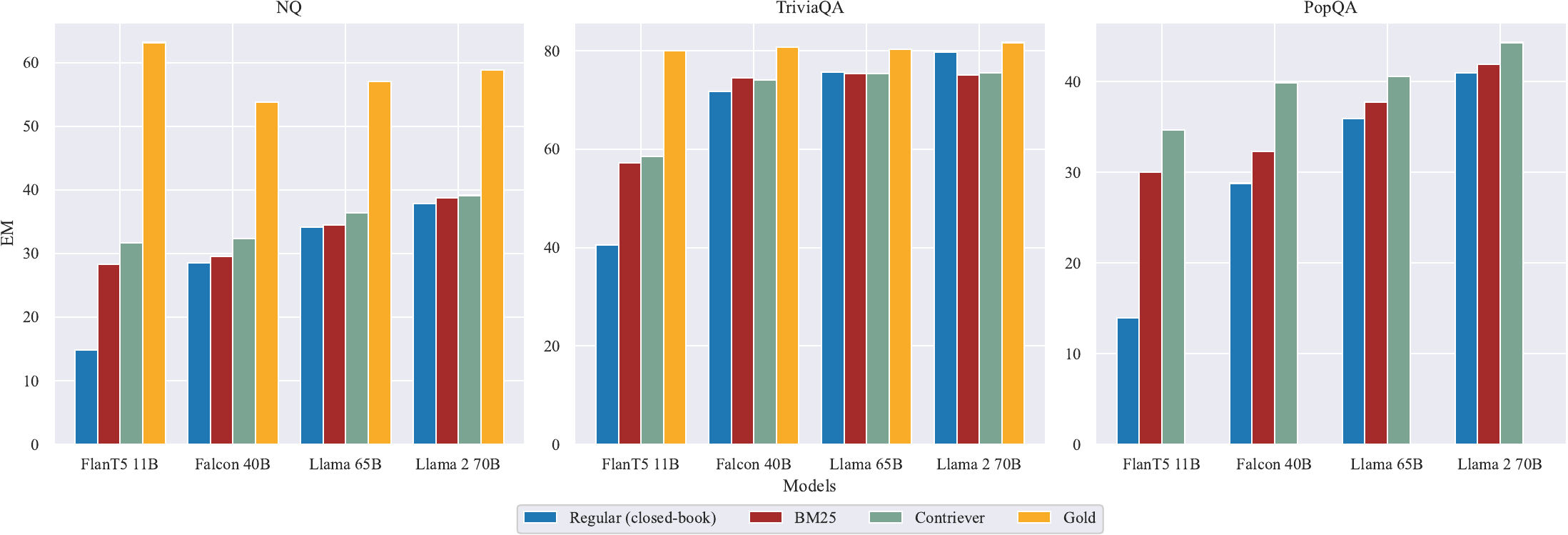}
\caption{\label{fig:retrieval-effect} Performance comparison between regular decoding and our method using different sources of retrievals.}
\end{figure*}

\subsection{Using Different Retrievals}
\label{sec:effect-of-context}
As previously highlighted, our investigation centers on retrieval-augmented LLMs, involving the implementation of retrieval modules over a knowledge base concerning a user query. Subsequently, the retrieved relevant passage supplements the prompt to facilitate the generation of answers by the LLM. In earlier experiments, we utilized the provided gold context by NQ and TriviaQA to establish the theoretical upper bound of our proposed decoding method. This segment aims to examine whether the utilization of off-the-shelf retrieval mechanisms would influence the efficacy of our proposed methods. In Figure~\ref{fig:retrieval-effect}, we present a comparative analysis between closed-book regular decoding and our decoding method, utilizing retrieval passages from BM25, Contriever, or the provided gold context.

It is pertinent to note that the PopQA dataset lacks gold context. The comparative analysis indicates that results derived from Contriever exhibit superiority over those derived from BM25. Moreover, a substantial disparity exists between outcomes obtained through retrieval and those derived from leveraging gold context. It is essential to underscore that while these observations do not negate the efficacy of our proposed decoding method, they do suggest that enhancements to the retrieval module could yield improved outcomes.

\begin{table}[ht]
\centering
\begin{tabular}{lccc}
\toprule
\textbf{Irr. Passage} & {\textbf{NQ}} &  {\textbf{TQA}} &  {\textbf{PopQA}} \\
                           \midrule
 Random              & 56.74 & 81.28 & 43.23     \\
 Fixed              & 57.95  & 80.84 & 43.82     \\
 Fixed (rand. perm.)    & 57.17 & 80.68 & 42.98     \\
 Most distant   & \textbf{58.86} & \textbf{81.7} & \textbf{44.3}   \\   

\bottomrule
\end{tabular}
\caption{Comparison of performance on Llama 2 70B across various methods for selecting irrelevant $\boldsymbol{c}^-$: random selection, fixed adversarially constructed contexts, fixed context with random word permutation, and passages with the most distance from the relevant context.}
\label{tab:irrelevant_context_results}
\end{table}

\subsection{Selection of Irrelevant Context}
\label{sec:effect-of-irrelevant-c}

An essential aspect of our decoding method involves the incorporation of the $\boldsymbol{c}^-$ irrelevant context. Here, we investigate various strategies for selecting $\boldsymbol{c}^-$ and its impact on our methods. Initially, we propose employing a random selection of $\boldsymbol{c}^-$ from the complete pool of available contexts (ensuring that the randomly selected $\boldsymbol{c}^-$ differs from $\boldsymbol{c}^+$). Subsequently, we manually construct an adversarial $\boldsymbol{c}^-$ devoid of meaningful or useful information, details of which are provided in Appendix~\ref{app:fixed_irrelevant_context}. Additionally, we experiment with shuffling the word order within this fixed $\boldsymbol{c}^-$. Another approach for determining $\boldsymbol{c}^-$ involves using lower-ranked retrievals. However, increasing the retrieval size arbitrarily is computationally inefficient, and even within the top-100 retrievals, relevant information can be present. Therefore, we approximate the bottom-ranked retrieval by selecting the $\boldsymbol{c}^-$ that exhibits the most distance from $\boldsymbol{c}^+$, based on the cosine distance of their embeddings in the retrieval module. The comparison results using Llama 2 70B are presented in Table~\ref{tab:irrelevant_context_results}. It is evident that $\boldsymbol{c}^-$ with the most distance yields the best performance. Throughout our experiments detailed in this study, if not explicitly specified, we employ the most distant option for selecting $\boldsymbol{c}^-$. However, if computing distance proves computationally expensive, the use of a fixed adversarial $\boldsymbol{c}^-$, as demonstrated in our results, remains a viable alternative.

\begin{figure*}[ht]
    \centering
    \includegraphics[width=\linewidth]{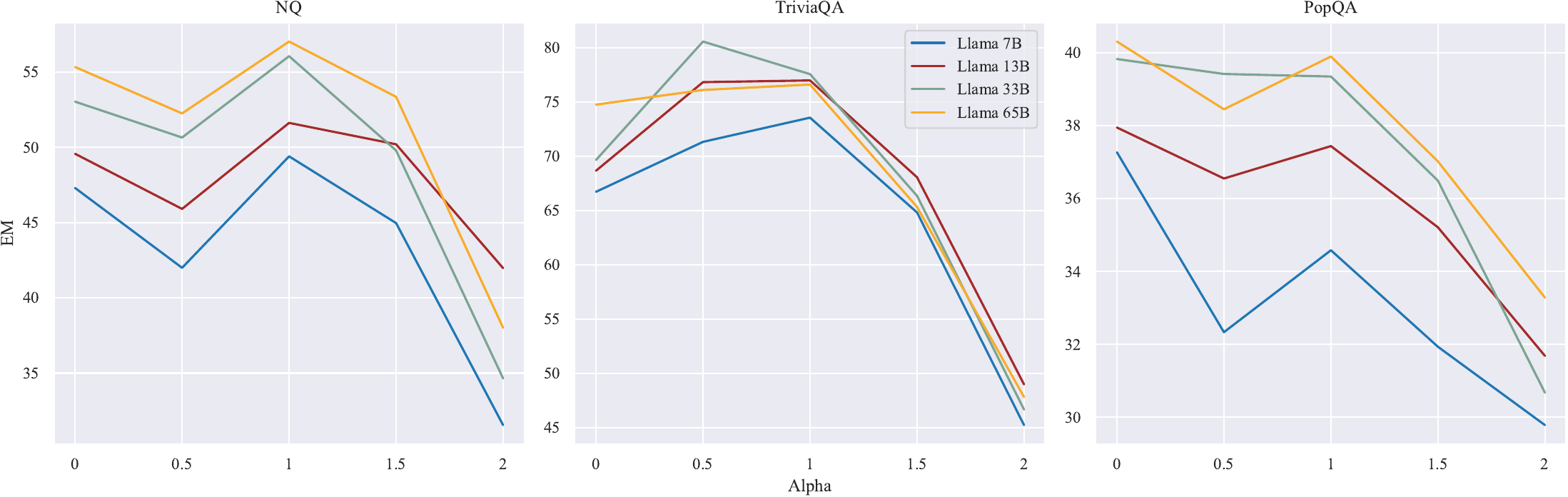}
\caption{\label{fig:alpha-effect}The impact of $\alpha$ on our decoding method across different sizes of Llama 1 models.}
\end{figure*}

\subsection{Adjusting the Knowledge Modification}
\label{sec:effect-of-alpha}

Our proposed decoding method introduces the hyperparameter $\alpha$, regulating the degree of modification applied to the parametric answer for a given input query. A larger $\alpha$ signifies a more substantial modification, while $\alpha=0$ denotes no alteration, thereby reducing decoding to a regular decoding scenario. Despite outlining a strategy to dynamically set this alpha value, we remain interested in assessing the impact of different alpha values on the efficacy of our method. We conducted experiments involving the adjustment of $\alpha$ levels and present the outcomes obtained from Llama models in Figure~\ref{fig:alpha-effect}. Our analysis reveals that as the alpha values increase, the effectiveness of the method diminishes substantially. The model achieves optimal performance at $\alpha=1.0$ , outperforming all other alpha settings. Furthermore, setting $\alpha=1.0$  yields consistent improvements over regular decoding on both the NQ and TriviaQA datasets. For PopQA, while a fixed $\alpha$ value offered no improvement over regular decoding, the dynamic setting we propose led to significant gains, as shown in Table~\ref{tab:main_results}.

\begin{figure}[t]
    \centering
    \includegraphics[width=\linewidth]{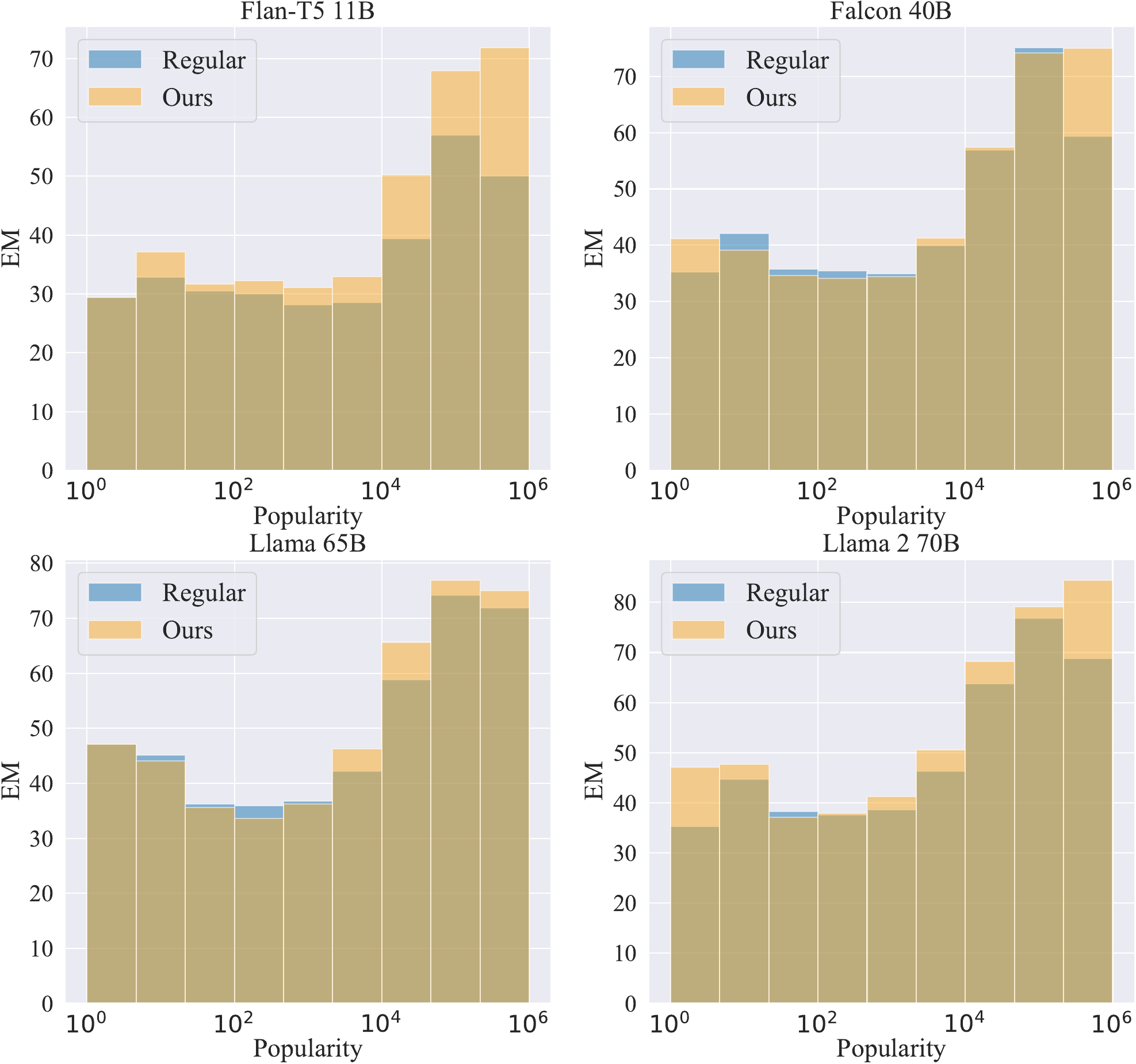}
\caption{\label{fig:popularity-effect}Comparison of performance between regular decoding (open-book) and our method on questions with varying levels of knowledge popularity.}
\end{figure}

\subsection{Answering across Knowledge Popularity}
The utility of retrieval mechanisms becomes evident in addressing less prevalent factual knowledge, an area where LLMs often exhibit limitations. Therefore, we conducted an analysis to evaluate the efficacy of our proposed decoding approach in facilitating LLMs to accurately respond to factual questions across a spectrum of popularity. Following \citet{mallen-etal-2023-trust}, we utilized the popularity of entities gauged by Wikipedia's monthly page views as an indicator of their frequency in web discussions. Our findings, presented in Figure~\ref{fig:popularity-effect}, juxtapose the performance of models employing regular decoding within an open-book setting against those employing our proposed method. The results manifest a consistent trend wherein our proposed method consistently outperforms regular decoding under an open-book setting across varying levels of popularity. This observation underscores the efficacy of our decoding strategy in assisting LLMs to generate more accurate responses to factual queries across a diverse range of entity popularities.

\begin{table}[ht]
\small
\centering
\begin{tabular}{llc}
\toprule
{\textbf{Model}}  & \textbf{Decoding} & {\textbf{NQ-SUB}} \\
                           \midrule
\multirow{5}{*}{Flan-T5 11B} & Reg.-Cl.  & 0.19       \\
                            & Reg.-Op.  & 56.4       \\
                            & CAD      & 51.9      \\
                            & Ours    & \textbf{57.55}    \\  
\midrule
\multirow{5}{*}{Falcon 40B} & Reg.-Cl.  & 0.13    \\
                         & Reg.-Op.  &  46.78     \\
                         & CAD      & 45.79     \\
                         & Ours     & \textbf{48.34}    \\  
\midrule
\multirow{5}{*}{Llama 65B} & Reg.-Cl.  & 0.08     \\
                        & Reg.-Op.  &  59.25     \\
                        & CAD      &  60.41       \\
                         & Ours     & \textbf{61.65}   \\  
\midrule
\multirow{5}{*}{Llama-2 70B} & Reg.-Cl.  &  0.02    \\
                           & Reg.-Op.  & 57.63    \\
                           & CAD      & 53.23      \\
                         & Ours     & \textbf{58.34}    \\  

\bottomrule
\end{tabular}
\caption{Comparison of decoding methods on the knowledge conflict dataset. Reg.-Cl. and Reg.-Op. denote regular decoding in closed-book and open-book settings.}
\label{tab:knowledge_conflict_results}
\end{table}

\subsection{Resolving Knowledge Conflicts}
As previously highlighted in the manuscript, tasks reliant on knowledge typically draw from two knowledge sources: parametric knowledge, acquired during training, and non-parametric knowledge, accessed via retrieval modules during inference. The issue of knowledge conflicts, wherein the contextual (non-parametric) information contradicts learned knowledge, has been formally addressed by \citet{longpre-etal-2021-entity} to understand how models utilize these dual sources of knowledge.

To generate question-answer pairs manifesting knowledge conflicts, we followed the methodology proposed by \citet{longpre-etal-2021-entity}. Initially, we identified questions in the NQ dataset that contained named entity answers. Subsequently, we obtained the relevant context for each question and replaced the gold answer entity in the context with a random entity. In this setup, an accurate LLM should produce the substituted entity as the answer when provided with the question and the modified context, disregarding its pre-learned parametric answer. This resulting dataset, termed NQ-SUB, was created for assessing models in scenarios involving knowledge conflicts. The performance results on NQ-SUB are presented in Table~\ref{tab:knowledge_conflict_results}. Remarkably, all models exhibited poor performance in the regular closed-book setting, given that the task requires the model to disregard its parametric knowledge. However, our proposed decoding method demonstrated superior performance compared to both regular decoding and CAD on this knowledge conflict task. The comparative results emphasize the effectiveness of our proposed decoding approach in addressing knowledge conflicts, particularly in scenarios where models encounter contradictions between their learned and contextual knowledge.

\section{Conclusion}
This study introduces a novel decoding strategy, employing contrastive decoding to incorporate relevant and irrelevant context. Through diverse experiments and analyses across datasets and model scales, our approach consistently outperforms regular decoding methods. Notably, it excels in managing knowledge conflicts, surpassing both regular decoding and CAD. Moreover, our exploration of retrieval sources underscores the need for refining these modules to enhance efficacy. The demonstration of the method's effectiveness in open-domain question answering also sets the stage for future research. The method's versatility suggests potential applications in various generative tasks, motivating our future exploration in tasks like summarization.

\section*{Limitations}
Our study acknowledges several limitations that warrant consideration. First, we acknowledge the restriction imposed by employing a singular prompt template. The computational complexity inherent in our method limited the scope of experiments conducted within this framework. However, this constraint was pivotal in maintaining consistency across our comparisons, ensuring the reliability and robustness of the obtained results despite the limitation in the number of explored templates.

Secondly, while our decoding method was specifically showcased in the context of question answering using greedy decoding, another limitation of this study is that we haven't explored its application to other generative tasks. It's essential to note that our approach is designed as a general decoding framework applicable to various generative tasks. Thus, expanding this work to other domains such as summarization and mitigating hallucination \citep{maynez-etal-2020-faithfulness,zhao-etal-2020-reducing,ji-etal-2023-hallucination} remains a promising avenue for future exploration.

Furthermore, it's imperative to recognize that the scalability and generalizability of our method across different problem domains and decoding strategies might present further challenges and considerations. Extending our investigation to encompass a broader array of prompt templates and decoding strategies (such as nucleus sampling; \citealp[]{holtzman-etal-2020-curious}) could potentially reveal nuanced insights into the adaptability and effectiveness of our proposed method.

Additionally, it is crucial to note that the decoding time required for our method is longer than regular decoding, approximately three times longer, owing to decoding using three logits distributions simultaneously. However, there exists potential for mitigating the time complexity by distributing the decoding of different distributions across multiple GPU machines, thereby enabling parallelization and potentially reducing the computational overhead. This approach might alleviate the time constraints associated with our decoding method, rendering it more feasible for applications requiring low decoding latency.

\section*{Acknowledgements}
We would like to thank Christos Christodoulopoulos, Marco Damonte, and Clara Vania for their insightful discussions that contributed to this work. We also appreciate the assistance of Yash Malik, Miguel Angel Rubio, and Mohammad Shahdad in setting up the computing environment for our experiments. Finally, we thank the anonymous reviewers for their helpful feedback, which helped us improve the clarity and quality of the paper.

\bibliography{anthology,custom}

\newpage

\appendix

\section{Additional Results on Scaling Experiments}
\label{app:scaling_results}
\begin{figure*}[t]
    \centering
    \includegraphics[width=\linewidth]{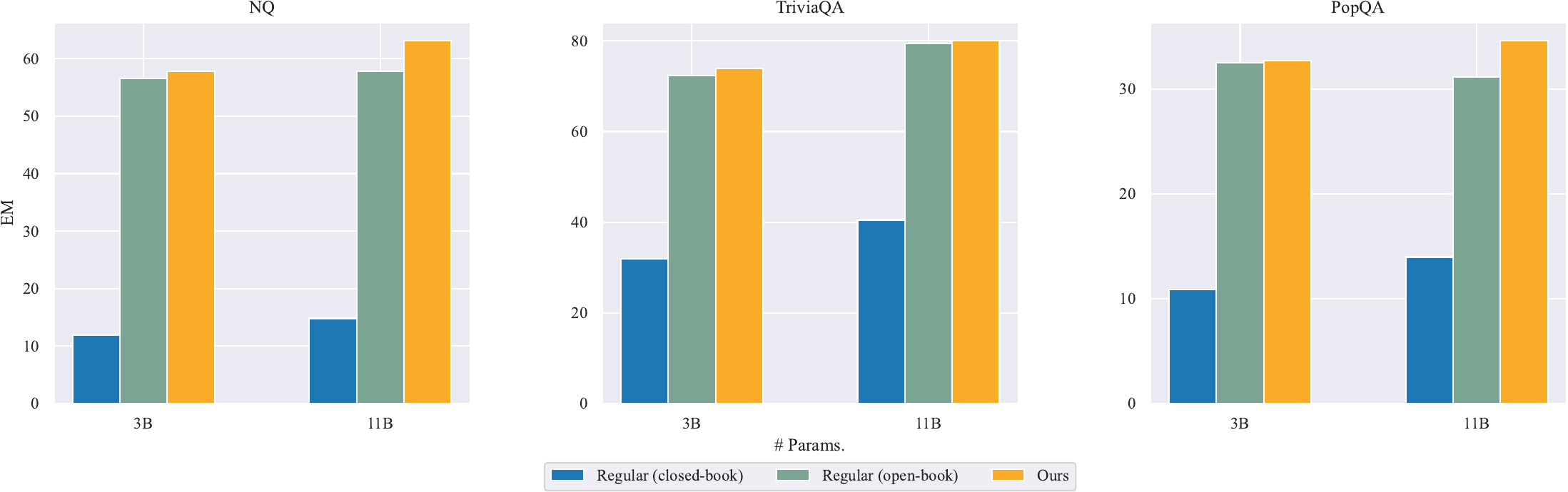}
\caption{\label{fig:scaling-effect-flan-t5} Performance comparison of our method against regular decoding across various sizes of Flan-T5 models.}
\end{figure*}

\begin{figure*}[t]
    \centering
    \includegraphics[width=\linewidth]{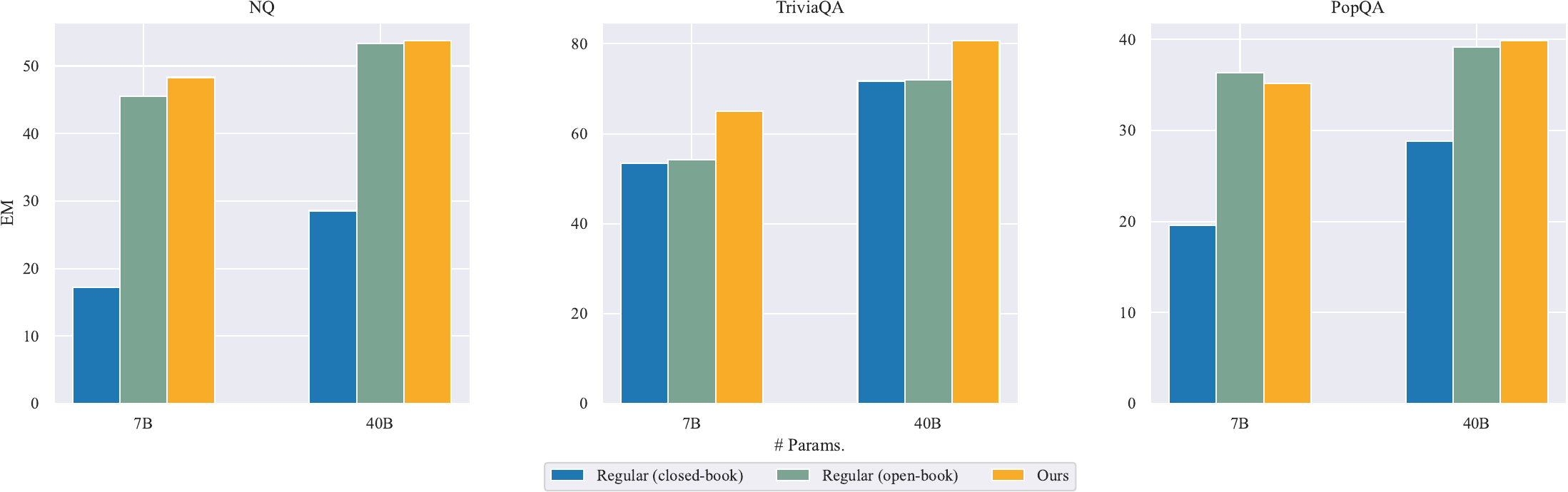}
\caption{\label{fig:scaling-effect-falcon} Performance comparison of our method against regular decoding across various sizes of Falcon models.}
\end{figure*}

\begin{figure*}[t]
    \centering
    \includegraphics[width=\linewidth]{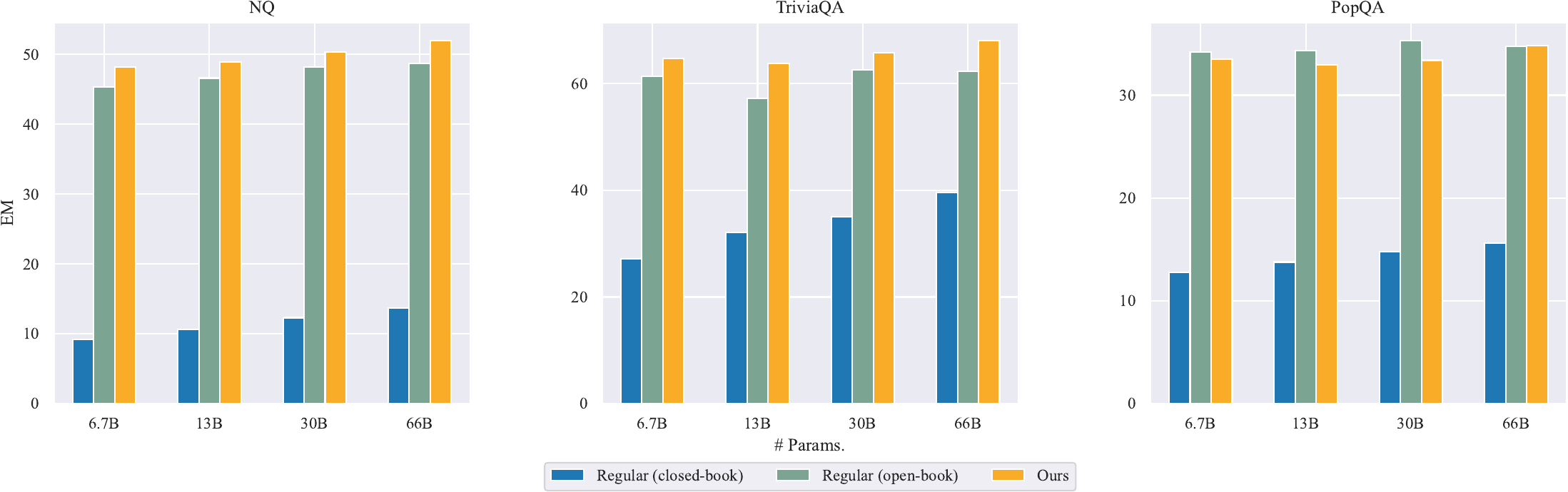}
\caption{\label{fig:scaling-effect-opt} Performance comparison of our method against regular decoding across various sizes of OPT models.}
\end{figure*}

\begin{figure*}[t]
    \centering
    \includegraphics[width=\linewidth]{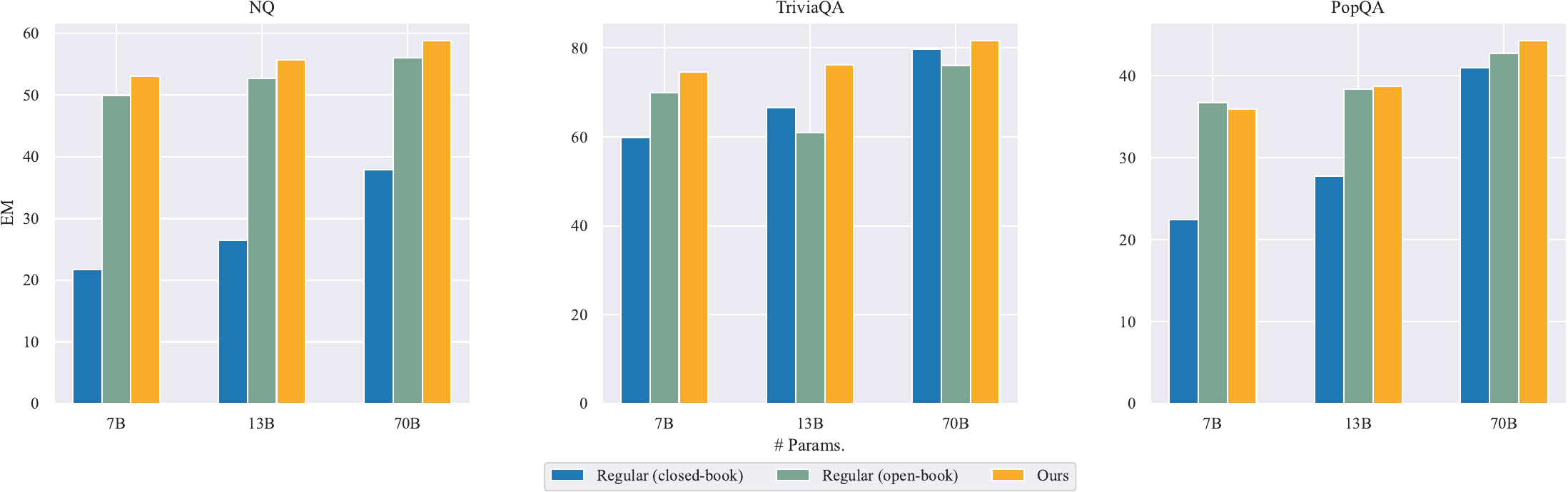}
\caption{\label{fig:scaling-effect-llama-2} Performance comparison of our method against regular decoding across various sizes of Llama 2 models.}
\end{figure*}

We present additional scaling experiment results for different model variants. Specifically, we illustrate the outcomes for Flan-T5 variants, 3B and 11B, in Figure~\ref{fig:scaling-effect-flan-t5}. The results for Falcon variants, particularly Falcon 7B and 40B, are depicted in Figure~\ref{fig:scaling-effect-falcon}. Moreover, we showcase the results for OPT variants, encompassing OPT 6.7B, 13B, 30B, and 66B, in Figure~\ref{fig:scaling-effect-opt}. Additionally, the findings pertaining to Llama 2 variants, including Llama 2 7B, 13B, and 70B, are illustrated in Figure~\ref{fig:scaling-effect-llama-2}. We can see that our proposed decoding method outperforms regular decoding with open-book setting in most settings across different datasets and model sizes.

\section{Additional Details on Irrelevant Context}
\label{app:fixed_irrelevant_context}
Here we provide the meticulously designed adversarial $\boldsymbol{c}^-$ irrelevant context that is used as the fixed $\boldsymbol{c}^-$ for every query:

\textit{``It was a pleasant weather day, with seasonally average temperatures. The local legislative and academic governing bodies held routine meetings regarding budgets and policies. Students focused on their studies while athletes practiced for upcoming competitions. Residents tended to their jobs and daily tasks around their neighborhood. Nothing particularly eventful occurred in the community. It was an ordinary midweek day. The weather was typical for the time of year without any extreme events. Overall it was an average day in the community with people pursuing their regular daily activities.''}

Here is the same fixed $\boldsymbol{c}^-$ but with word order permuted:

\textit{``an routine Overall was of community. average focused for The around tended upcoming their was policies. their budgets and Residents to eventful held competitions. It particularly extreme with academic temperatures. was day. weather local The their studies events. it meetings average pleasant typical Nothing ordinary time seasonally legislative people an the daily the Students in a neighborhood. activities. community pursuing weather and while in midweek regarding athletes occurred tasks the daily jobs It governing year bodies regular with their for day and practiced on day, was without any''}

\end{document}